\documentclass[final]{clv2025}

\jvol{vv}
\jnum{nn}
\jyear{2026}

\dochead{Position Paper} 

\pageonefooter{Action editor: Xuanjing Huang. Submission received: 12 June 2025; revised version received: 25 January 2026; accepted for publication: 04 March 2026.}

\usepackage{amsmath}
\usepackage{booktabs}

\usepackage{xcolor}
\usepackage{lipsum}
\usepackage{tabularx}
\usepackage{tablefootnote}
\usepackage[normalem]{ulem}
\newcommand{\mybox}[1]{\fcolorbox{black!30}{gray!10}{\scriptsize #1}}
\usepackage{graphicx}
\usepackage{diagbox}

\theoremstyle{break}

\def\axiomname{Axiom}

\usepackage{tikz}
\usetikzlibrary{arrows.meta}

\tikzset{
  nodestyle/.style = {circle, draw=black, align=center},
  edgestyle/.style = {->, >=Stealth}, 
  graphstyle/.style = {
    node distance = 1.5cm,
    every node/.style = {nodestyle},
    every path/.style = {edgestyle}
  }
}

\runningtitle{The Conundrum of Research on Attacking PII}
\runningauthor{Ochs \& Habernal}

\begin{document}

\title{The Conundrum of Trustworthy Research on Attacking Personally Identifiable Information Removal Techniques}

\author{Sebastian Ochs$^{1, 2}$\thanks{Corresponding author}, Ivan Habernal$^{1, 3}$}

\affilblock{
    \affil{Trustworthy Human Language Technologies\\ \quad \url{www.trusthlt.org}}
    \affil{Technical University of Darmstadt, Germany\\ \quad \email{sebastian.ochs@tu-darmstadt.de}}
    \affil{Research Center for Trustworthy Data Science and Security, \\ \quad Ruhr University Bochum, Germany \\ \quad \email{ivan.habernal@ruhr-uni-bochum.de}}
}

\maketitle

\begin{abstract}
Removing personally identifiable information (PII) from texts is necessary to comply with various data protection regulations and to enable data sharing without compromising privacy.
However, recent works show that documents sanitized by PII-removal techniques are vulnerable to reconstruction attacks.
Yet, we suspect that the reported success of these attacks is largely overestimated.
We critically analyze the evaluation of existing attacks and find that data leakage and data contamination are not properly mitigated, leaving the question whether or not PII removal techniques truly protect privacy in real-world scenarios unaddressed.
We investigate possible data sources and attack setups that avoid data leakage and conclude that only truly private data can allow us to objectively evaluate vulnerabilities in PII removal techniques.
However, access to private data is heavily restricted---and for good reasons---which also means that the public research community cannot address this problem in a transparent, reproducible, and trustworthy manner.
\end{abstract}

\section{Introduction}
Nobody wants their private text data, such as personal messages, medical notes, or presence in court rulings, to be freely accessible by the general public.
Simultaneously, research on sensitive text domains like law and healthcare is necessary for developing technologies that benefit humanity \cite{DEMNERFUSHMAN2009760, Sudlow2015, zhong-etal-2020-nlp}.
To conduct research on private text data while complying with regulations and privacy laws, personally identifiable information (PII) must be removed from affected documents.
Manually removing PII from texts is a slow and tedious process \cite{Dorr2006-kd} that does not scale well given the amounts of sensitive documents available in certain domains \cite{faucris.272404594}.
PII removal tools \cite{Sweeney1996-iz, textwash} aim to protect the privacy of individuals to comply with the legal frameworks, while enabling publishing useful data that no longer contains sensitive information.
However, PII removal tools do not guarantee privacy in any formal way.
Instead, they mimic how humans usually process text documents to identify and remove private information from the data.
This leaves the PII removed texts vulnerable to attacks, as humans can partially reconstruct the removed information with significant effort and background knowledge \cite{carrell2020resilience, 285449}.
Meanwhile, large language models (LLMs) are becoming increasingly more powerful and are capable of inferring private information from text \cite{staab2024beyond}.
This raises concerns about whether LLMs may also be able to infer sensitive information from PII removed documents.

Existing approaches that attack PII removed texts claim that they successfully recover parts of the private information \cite{nyffenegger-etal-2024-anonymity, Patsakis2023-de, 10179300, charpentier-lison-2025-identification}.
However, it remains unclear whether their attacks succeed because the PII removal method itself is insufficient, or because the simulated attacker already has access to the original data, for example through applying the attack on well-known public data or by data contamination and data memorization by the attack model.
Moreover, there is a lack of guidance on how to properly design, execute, and evaluate adversarial attacks against PII removal tools.

We therefore pose two research questions.
First, is the evaluation of existing methods for PII reconstruction attacks fundamentally flawed?
In this paper, we show what contributes to the inflation of re-identification scores and overestimation of the general ability of the proposed attacks.
Second, is it possible for public researchers to address these potential flaws without access to real, sensitive data?
We argue that without access to private data, we cannot reliably determine whether the attack models used in previous research with undisclosed pretraining data, especially proprietary LLMs, are genuinely inferring sensitive information from PII removed texts or simply reproducing text spans they memorized during pretraining.
To highlight the current shortcomings of PII reconstruction attacks, we further conduct and analyze two experiments on real-world data from Czech court announcements (legal domain) and from personal English travel videos (YouTube `vlogs'), both extremely unlikely being leaked to any dataset for LLM pre-training.

In the following, we present and briefly discuss related work (Section~\ref{sec:rw}), and then provide background on PII in texts, regulations and its removal techniques (Section~\ref{sec:pii}).
Before addressing our research questions, we point out flaws in PII-related regulations and removal tools (Section~\ref{sec:pii-flaws}).
We then answer our first research question (Section~\ref{sec:crit}) and turn to the second research question by outlining a valid attack setup (Section~\ref{sec:setup}) and why real private data is needed for proper evaluation (Section~\ref{sec:data-discussion}).

\section{Related work}
\label{sec:rw}

While most of the related work deals with PII removal techniques or attacks, we are not aware of any existing work that directly questions whether the experimental design is inherently flawed.
For example, some surveys in the medical domain review the application of PII removal techniques to electronic health records from the US before the deep learning era \cite{Meystre2010, ford2016extract}.
Recently, \citet{Sousa2023} categorize several privacy-preserving NLP methods based on deep learning, including PII removal, and discuss their utility to comply with data protection laws. 

\citet{lison-etal-2021-anonymisation} identify several issues inherent to PII removal techniques (which they term `text anonymization').
The authors highlight that removing predefined categories of PII from text documents is insufficient to provide any formal privacy guarantees, and that human annotators often disagree about which text spans contain private information.
Furthermore, \citet{lison-etal-2021-anonymisation} compare typical PII removal methods based on named entity recognition to privacy-preserving data publishing methods from other domains, such as k-anonymity \cite{sweeney2002k}, C-sanitize \cite{sanchez2016c} and differential privacy \cite{10.1007/11787006_1} and discuss possible difficulties when applying these methods to text.
The authors argue against solely relying on PII removal techniques and emphasize the need to incorporate context, utility, and re-identification metrics into the private data-sharing pipeline.  
However, they do not question the potential flaws in attacks against PII removal techniques or investigate the role of information leakage in these attacks.

In this paper, we therefore not only argue that there exist flaws in PII removal techniques, but there are also flaws in the attacks against them.

\section{Personally identifiable information in texts}
\label{sec:pii}

Personally identifiable information (PII) refers to any attributes that can be used---either alone or in combination---to uniquely identify a person.
Data, especially documents from the medical, legal, or financial domain, often contain names, addresses or phone numbers that can single out an individual on their own, called \emph{direct identifiers}.  
Besides direct, there also exist \emph{indirect identifiers} or \emph{quasi-identifiers}, which can distinguish a person when they are combined with each other.
For example, date of birth, ZIP codes, gender or occupation are not unique to a person, but linking databases together where those attributes appear may lead to the identification of individuals \cite{37926}.
In contrast to traditional databases, natural language texts can also reveal personal information through syntactic features, such as writing style \cite{koppel2002automatically, verhoeven-etal-2016-twisty}, or semantic content that can be linked to PII \cite{elazar-goldberg-2018-adversarial, staab2024beyond}.

\subsection{PII in regulations}
When a dataset is shared with third parties, e.g. for data analysis or research, it is essential that the privacy of contributing persons is protected.
Therefore, legislators around the world created different legal frameworks for handling private data.
While these regulations often establish similar definitions of PII and require the removal of both direct and indirect identifiers from texts, they differ in the scope of what must be removed or conditions under which some identifiers may be retained. 

The General Data Protection Regulation (GDPR, \citet{gdpr}), for example, describes how personal data (synonymous with PII) of individuals in the EU must be protected. 
It lists several types of personal data, particularly the ``name, an identification number, location data, an online identifier'' and ``factors specific to the physical, physiological, genetic, mental, economic, cultural or social identity'' of a person (GDPR, Art.~4(1), \citet{gdpr}). 
Additionally, the GDPR also defines categories that require stricter protection, such as the ``racial or ethnic origin, political opinions, religious or philosophical beliefs'' of an individual (GDPR, Art.~9(1), \citet{gdpr}).

In contrast to the GDPR's broad coverage, the U.S. Health Insurance Portability and Accountability Act (HIPAA, \citet{hipaa}) applies more specifically to personal information found in medical data. 
It defines the `Safe Harbor' method, which requires the removal of 18 specific types of protected health information (PHI: including names, location and phone numbers) for medical documents to be considered de-identified.
Once the method is applied to health-related documents and all PHI are removed, the data is no longer subject to the HIPAA privacy rules and may be published.

Similar to how HIPAA regulates removing PHI of medical data in the U.S., the GDPR serves as a legal framework to ensure the protection of individuals' privacy when publishing documents. 
For example, decisions of the `Court of Justice of the European Union' are `anonymized' by replacing names of natural persons with synthetically generated, fictitious names and released to the public, (apparently) meeting GDPR standards.\footnote{See ``Fictitious names in anonymised cases'' at \url{https://curia.europa.eu/jcms/jcms/p1_3869098/en/}}
In recent years, algorithmic tools are increasingly applied to remove PII from court documents, which speeds up the publication process while still complying with GDPR privacy protection across EU jurisdictions \cite{10.1145/3594536.3595151}.
However, the implications of taking the GDPR seriously for PII removal and NLP in general can be disastrous. \citet{weitzenboeck2022the} show that if one wants make sure text data are anonymized according to the most stringent interpretation of the GDPR, the redacted text output becomes just useless \citep[Table~2 on p.~197]{weitzenboeck2022the}. This stringent interpretation of the GDPR should prevent any linkage attacks; in other words, it must be impossible to link the anonymized documents to any other documents that contain personal information. But here we encounter an unsolvable problem: \citet{weitzenboeck2022the} anonymize court documents against an adversary \emph{who already owns the original unredacted documents}. Therefore, as long as someone has a legal obligation to keep the original document, which is the case for many datasets, publishing a useful anonymized version violates the GDPR on the one hand, but a strict anonymization is useless on the other hand.

\subsection{Definitions}
\begin{table*}[t]
\centering
\small
\begin{tabularx}{\textwidth}{@{}lX@{}}
\toprule
\textbf{Technique} & \textbf{Example} \\
\midrule
\textbf{Original} & Hi, I'm Jane Doe, and I live at 123 Maple Street, Springfield. You can reach me at (555) 123-4567 or email me at janedoe@email.com. \\
\addlinespace
\textbf{Redaction} & Hi, I'm , and I live at 123 , . You can reach me at  or email me at . \\
\addlinespace
\textbf{Masking} & Hi, I'm ********, and I live at 123 ************, ***********. You can reach me at ************** or email me at ***************om. \\
\addlinespace
\textbf{Replacement (Tag)} & Hi, I'm <PERSON>, and I live at 123 <LOCATION>, <LOCATION>. You can reach me at <PHONE\_NUMBER> or email me at <EMAIL\_ADDRESS>. \\
\addlinespace
\textbf{Pseudonymization} & Hi, my name is Alex Carter, and I live at 123 Oak Drive, Rivertown. You can reach me at (206) 482-7743 or via email at acarter88@mailnet.com. \\
\bottomrule
\end{tabularx}
\caption{We generated a message from a ficticious person with ChatGPT 4o mini (\url{https://chatgpt.com/share/681de483-181c-800e-ae40-88cc2fefb0f9}) and applied common PII removal techniques to the same input sentence with Microsoft Presidio (\url{https://huggingface.co/spaces/presidio/presidio\_demo}).}
\label{tab:pii-removal}
\end{table*}

In related works, the terminology associated with the removal of PII is often ambiguous.
We therefore define several key terms to clarify their usage in the context of natural language processing in our paper.

\paragraph{PII Removal} We use \emph{PII removal} as a general term for the practice of removing a predefined set of private information from texts to mitigate privacy risks. 

\paragraph{Anonymization} \emph{Anonymization} is the process of removing or modifying PII from text documents, such that re-identification is not possible with current technology.
As stated in the GDPR, data is considered \emph{anonymized} if re-identification of a particular person can not be reasonably achieved by linking the data to other datasets or applying any currently existing adversarial attacks at the time of the data processing~\cite[Recital 26]{gdpr}. 
It should be noted that this definition does not account for potential future attacks.

\paragraph{De-identification} According to HIPAA, \emph{de-identification} involves altering or removing PII from text documents so that either re-identification risks are low, according to a human expert, or all occurrences of the 18 designated PHI are removed \cite{cfr_45}.
In contrast to anonymization, \emph{de-identification} is therefore less strict as HIPAA accepts a low level of risk, if the effort required for re-identification is deemed sufficiently high by the expert's judgment. 
For example, the electronic health records aggregated by CancerLinQ are de-identified using expert determination \cite[p.~3]{Potter2020cancer}. 

\paragraph{Sanitization} Text \emph{sanitizing} describes the process of obscuring sensitive information in texts while preserving their utility \cite{wei2009tplaus, papadopoulou-etal-2022-neural, olstad-etal-2023-generation}.
The privacy protection mechanisms of \emph{sanitization} are not necessarily limited to removing or modifying PII in texts, but may also be based on differentially private rephrasing methods \cite{yue-etal-2021-differential, chen-etal-2023-customized}.
Unlike anonymization or de-identification, \emph{sanitization} does not follow a unified re-identification risk standard like HIPAA or GDPR, but may utilize existing frameworks.

\subsection{Typology of PII removal techniques}
\paragraph{Redaction}    Deleting PII from text documents is referred to as \emph{redaction}. 
This process can disrupt the syntactic coherence of sentences as grammatical constituents such as names or other entities are completely removed.

\paragraph{Masking} {Masking} obscures PII by substituting them with non-semantic placeholders, for example mask tokens. 
These placeholders reveal the placements of PII in the original texts but do not disclose any information beyond their position \cite[Table~2]{berman2003concept}.

\paragraph{Replacement} Preserving the contextual meaning of sensitive data is achieved by \emph{replacing} PII with more relevant but general information. 
For example, names could be replaced by a special token [name] or locations with [location] \cite[Table~3.5]{neamatullah2008automated}.

\paragraph{Pseudonymization} 
The GDPR defines \emph{pseudonymization} as replacing PII with substitutes so that it can not be connected to the affected person without the use of additional data \cite[Art. 4 (5)]{gdpr}.
Under this definition, the relation between the PII and their surrogates can be stored to enable later reconstruction of the original data.
However, the primary goals of \emph{pseudonymization} in NLP are privacy preservation and utility, not the ability to reconstruct the original texts.
We therefore deviate from the GDPR in this paper and describe \emph{pseudonymization} as substituting PII with synthetically generated data without keeping the link between the synthetic and original information.
In NLP, \emph{pseudonymization} is often supported by context-aware language models to retain text utility for downstream tasks while privacy is preserved \cite{eder-etal-2022-beste}.\\

We provide examples of each described PII removal technique applied to a sentence containing ficticious PII in Table \ref{tab:pii-removal}. 

\subsection{PII removal tools}
The shift towards automatic PII removal from texts came in response to earlier practices.
Before automatization, PII was typically removed by manually blacking out their occurrences in text documents, a task that requires a lot of effort and is time-consuming \cite{Dorr2006-kd}.
Even nowadays, PII are manually removed from German court decisions, resulting in the publication of only 2\% of the rulings issued by German courts \cite{faucris.272404594}.  

\paragraph{Early methods}
One of the earliest tools was the Scrub system \cite{Sweeney1996-iz}, which used a rule-based approach to detect sensitive information and replaced it with handcrafted values from a lookup table.
Through pattern matching and handcrafted rules, Scrub replaced specific types of PII such as names, addresses, dates, and phone numbers from medical discharge summaries.
However, Scrub struggled with context-dependent entities or ambiguous terms, for example PII with references to family relationships or medical conditions.
In contrast, systems based on named entity recognition (NER), such as HIDE \cite{hide}, were reported to be more accurate and flexible in detecting a wider range of private information.
Rule-based techniques still play a significant roll in NLP to remove PII from text corpora, especially number-based information such as phone and credit card numbers or IP addresses, see \citet[Section~3.3]{laurenon2022the} or \citet[Section~5.3]{soldaini-etal-2024-dolma}.

\paragraph{Contemporary methods}
Currently, a popular open-source text de-identification tool is Presidio,\footnote{\url{https://github.com/microsoft/presidio}} a software which combines both rule-based and NER techniques to detect and replace PII.
Presidio offers support for multiple languages and different PII replacement methods, as displayed in Table \ref{tab:pii-removal}, and \citet{KOTEVSKI2022104880} suggest its use for de-identifying medical data.
One drawback of Presidio is lacking coreference resolution, which can lead to inconsistent de-identification when one piece of private information is referenced multiple times and in different forms within a document.
Other methods like Textwash \cite{textwash} employ coreference resolution, but only for specific PII types that are easily traceable, such as names and locations. 
Beyond using NER and rule-based systems, \citet{liu2023deidgptzeroshotmedicaltext} propose to de-identify texts with GPT-4 in a zero-shot setting. 
While they achieve high accuracy on a clinical notes dataset, it is questionable whether private information can be sent to an online LLM provider in the first place, as it inadvertently leaks the sensitive data. 

\paragraph{Benefits of PII removal tools}
Automated PII removal tools are very efficient at removing well-structured sensitive information from free-form texts.
With the integration of more sophisticated detection models based on NER, their performance on context-dependent PII has significantly improved compared to earlier rule-based approaches \cite{pilan-etal-2022-text}.    
PII removal seems to offer an empirically sound defense against attacks, as it currently takes significant effort and extensive background knowledge to re-identify persons from de-identified texts \cite{carrell2020resilience, 285449}. 
Tools such as Presidio and Textwash also support data holders to comply with the privacy protection standards defined in legal frameworks.

\subsection{Defense methods beyond PII removal}

A key limitation of PII removal is the degradation of text utility. By redacting all explicit identifiers, the resulting text often loses key features that are essential for meaningful data analysis \cite{pal2024the}.
While Presidio and Textwash offer different strategies to replace PII with semantic placeholders (pseudonymization), researchers also explored other options such as text rewriting to better preserve the semantic meaning of documents while removing PII.

\paragraph{Adversarial text rewriting}

\citet{staab2025language} propose to iteratively rewrite a text document with GPT-4 to obscure the sensitive information detected by an adversarial GPT-4 model. \citet{yang2024robustutilitypreservingtextanonymization} extend this approach by also including an LLM aiming to preserve the utility of the rewritten text for downstream tasks. 
These methods, however, raise some privacy concerns, as the data is sent to the closed-source GPT-4 model for inference, inevitably sharing the private data with a closed-source company before the sanitization process.
Even if adversarial text rewriting methods empirically defend stronger against the selected adversaries in the respective paper than PII removal techniques, they still lack any formal privacy guarantees.
Given these limitations, and since our main focus are PII reconstruction attacks, we did not discuss adversarial text rewriting methods further in our paper.

\paragraph{PII removal combined with DP}
One hybrid approach that combines PII removal with differential privacy is presented by \citet{mouhammad-etal-2023-crowdsourcing}. 
The authors first detect PII within texts via an NER-based approach, replace the identified spans with special tokens, and then rewrite the resulting text under differential privacy guarantees \cite{igamberdiev-etal-2022-dp}.
However, their results show that text utility suffers considerably, even when using very high values of $\epsilon$ that offer little to no meaningful privacy protection.
We therefore did not consider such approaches for further discussion in our position paper.

\section{PII-related privacy regulations and removal tools have inherent flaws}
\label{sec:pii-flaws}
Despite their advantages, PII regulations and removal techniques have limitations that can prevent them from fully protecting the privacy of individuals. 

Interpretations of what exactly constitutes personal data differ across legal frameworks from various countries.
This poses a challenge to the developers of PII removal tools, as complying with one regulation does not guarantee  
compliance with others.
In the US, for example, text de-identification systems built for HIPAA focus on removing the 18 types of PHI from medical documents to adequately protect patient privacy in accordance with the legislation.
But when these tools are then applied to other domains, such as financial data, they may fail to detect unique identifiers from other domains , e.g., financial codes like IBANs or SWIFT/BIC codes, which fall outside of the scope of HIPAA.
Beyond domain-dependency, many countries---such as China and Brazil---have adopted privacy laws inspired by the GDPR.\footnote{\url{https://iapp.org/news/a/three-years-in-gdpr-highlights-privacy-in-global-landscape}}
However, the subtle differences between different regulations complicate the generalization of PII removal tools beyond the adaptation to a new language.
For example, the GDPR explicitly lists sexual orientation and political beliefs as sensitive personal information \cite[Art.~9(1)]{gdpr}, while the Chinese Personal Information Protection Law (PIPL) does not.\footnote{\url{https://personalinformationprotectionlaw.com/PIPL/article-28/}}

\paragraph{Unbounded list of PII}
More challenging is the fact that the list of PII requiring removal from texts is potentially unbounded, as personal data is broadly interpreted as ``any information relating to an identified or identifiable natural person'' according to Art.~4(1) of the GDPR \cite{gdpr}.
While the regulation provides some examples of personal data, it deliberately leaves the set of PII open-ended.
This significantly increases the difficulty for data holders to share their data with the research community, as the degree of safeguarding required by the GDPR \cite[Art.~89]{gdpr} is undefined \cite{Peloquin2020-gdpr, Vlahou2021-og}.
In the text domain, successful anonymization is therefore subject to interpretation by the legislators, with data holders facing the risk of non-compliance, trial in courts and potential fines when personal data is exposed after anonymized release.\footnote{\url{https://openjur.de/i/openjur\_wird\_verklagt.html}}
Given a strict interpretation of WP 216\footnote{\url{https://ec.europa.eu/justice/article-29/documentation/opinion-recommendation/files/2014/wp216\_en.pdf}} (a guidance document on anonymization issued by EU regulators before the GDPR, which remains influential), \citet{weitzenboeck2022the} even argue that anonymization of unstructured data is essentially impossible.

\paragraph{Vague capabilities of the attacker}
In cryptography, encryption methods are designed by considering future resources of the attacker to ensure the protection of data \cite[Section~VI]{1055638}. 
For example, existing research on cryptographic algorithms is motivated by the future threat to the existing encryption standards posed by quantum computing.\footnote{\url{https://csrc.nist.gov/projects/post-quantum-cryptography}}
In contrast to cryptography, where future advancements are considered, the capabilities of the attacker on texts protected by PII removal techniques are not sufficiently defined by regulations when it comes to protecting privacy in textual data.
While the GDPR states that current state of the art re-identification methods and their technological development have to be taken into consideration at the time of the text anonymization process~\cite[Recital 26]{gdpr}, it does not define the resources of the attacker beyond the criterion of ``means reasonably likely to be used'' for re-identification.
This makes it difficult for data holders to adequately address vulnerabilities emerging from the advancements in NLP and LLMs, which threaten the privacy of the individual, such as training data extraction \cite{274574} or LLM inferences of private information \cite{staab2024beyond}.
Without a clear attack formalization to follow against text anonymization, researchers improvise their own attack setups, potentially leading to flawed attack scenarios, a topic we explore in more detail in Section~\ref{sec:crit}.

\paragraph{Lack of guarantees against future attacks}
The major drawback of PII removal techniques, however, is that they do not provide formal privacy guarantees that would be impossible to overcome, under the formally-specified technique and the explicitly stated attacker capabilities.
In contrast to differential privacy \cite{10.1007/11787006_1}, which offers mathematically provable (yet probabilistic) privacy protection for data, \textbf{the privacy provided by PII removal is solely based on empirical risk evaluations that can not account for potentially stronger attack methods}.
  
\section{Critical analysis of adversarial attacks on data protected by PII removal}
\label{sec:crit}
\begin{table*}[t]
\centering
\small
\begin{tabularx}{\textwidth}{@{}lX@{}}
\toprule
\textbf{System} & Find which celebrity matches the following description and
match the missing information \\
\midrule
\textbf{User} &  PERSON\_FIRSTNAME\_1 PERSON\_LASTNAME\_2 is a NUMERIC\_4 year old actress, born in LOCATION\_2 but raised in LOCATION\_1. PRONOUN is most famously known for playing PERSON\_FIRSTNAME\_2 PERSON\_LASTNAME\_1 in the OTHER\_IDENTIFYING\_ATTRIBUTE\_2 franchise. [...] PRONOUN is very involved in activism and is a very outspoken feminist.\\
\midrule
\textbf{Assistant} & \mybox{Emma} \mybox{Watson} is a \mybox{31} year old actress, born in \mybox{Paris} but raised in the \mybox{UK}. She is most famously known for playing \mybox{Hermione} \mybox{Granger} in the \mybox{Harry Potter} franchise. [...] \mybox{she} is very involved in activism and is a very outspoken feminist.\\
\bottomrule
\end{tabularx}
\caption{Shortened example from Table 5, \citet{Patsakis2023-de} of GPT-4 recognizing Emma Watson from a PII removed text (\url{https://chatgpt.com/share/6825cf22-06e4-800e-9de1-edf923c9dc23}).
}
\label{tab:emma}
\end{table*}

Early attempts at de-identifying structured data containing textual information led to inadequate privacy protection, which resulted in the exposure of personal information of individuals who provided their data to platforms, certainly under the assumption that it would remain anonymous \cite{4531148, Sweeney2017-re}.\footnote{\url{https://www.cnet.com/tech/tech-industry/aol-apologizes-for-release-of-user-search-data/}}
Such data leaks may have influenced the decision of legislators to implement stricter privacy regulations to protect user data, such as HIPAA or the GDPR.
Current state-of-the-art de-identification methods therefore remove more PII from texts than required by previously implemented privacy regulations. 
But with the growing reasoning capabilities of LLMs \cite{mao-EtAl:2024:main}, redacted PII might be at risk to be revealed through attacks.
However, \textbf{are these attacks successful due to stronger model abilities, such as inferences from context, or due to inadequate attack setups and potential side-channel leaks?}

To address our first research question, whether existing evaluations of PII reconstruction attacks are fundamentally flawed, we focus on recent works that explicitly target documents where PII removal was applied and LLMs are the main attack method, either through retrieval-augmented generation, direct prompting or infilling.
To identify such publications, we resorted to an extensive manual search, as systematic, keyword-based searches proved to be ineffective, due to the small yet fragmented research area, where terminology is used inconsistently across papers. 
While this selection is not exhaustive, it is representative of the current directions in adversarial attacks on PII-removed texts with LLMs.

\subsection{Leakage through media reporting}
\label{sec:leakage.through.media.reporting}

We exemplify a particular type of side-channel leakage using the following paper.
\citet{nyffenegger-etal-2024-anonymity} proposed an attack on PII-sanitized texts, namely on court rulings of the federal supreme court of Switzerland, which were sanitized by masking PII through human annotators assisted by an automatic PII detection system.
The authors manually re-identified persons involved in seven court decisions to create a gold standard for the attack by manually linking the rulings to Swiss news articles that contain relevant information about each case, for example utilizing the penalty or case file numbers mentioned in the media. 
Using the additional information, the researchers link more news articles to the case until they successfully re-identify the name of seven individuals that were involved in the court rulings.
For the final dataset, the researchers also mix in irrelevant articles to ensure that the resulting dataset does not only include relevant data for the court decisions. 
The attack itself is an approach based on retrieval-augmented generation (RAG) with LLMs with the goal of recovering the name of the case participants.
The authors retrieve the top five relevant articles for each sanitized court ruling and prompt GPT-4 to re-identify the person from the given context, which was successful for 5 out of the 7 cases.

The attack framework presented by \citet{nyffenegger-etal-2024-anonymity} does not suffice to disprove the effectiveness of PII removal methods.
As the authors report, when providing GPT-4 with relevant news articles that contain the real name of a litigant in the sanitized court ruling, the model is able to recognize the person.
However, in these cases, the news articles reporting on the court rulings are already public knowledge and are the decisive factor that lead to the re-identification of affected parties.
\textbf{PII removal techniques can not offer protection in such cases, as the underlying data is already publicly available information before the text sanitization, and the names of the case participants are no longer private.}

\subsection{Leakage through public knowledge}
Another attack on sanitized text documents is proposed by \citet{Patsakis2023-de}.
For their attack, the authors utilize a dataset published by Textwash that contains manually rewritten Wikipedia biographies of celebrities.
Textwash replaced the PII in the biographies with corresponding named entity tags. 
The researchers attacked the sanitized biographies with the aim to recognize the original celebrity. 
Using GPT-3 \cite{NEURIPS2020_1457c0d6}, the attack successfully recognized 59\% of famous people.
An alternatively tested strategy using PII with pseudonymization to intentionally mislead the attacker led to a drop in re-identification, down to 50\%, after revising the attack prompt.

However, a major limitation of the proposed attack by \citet{Patsakis2023-de} is that it was conducted on well-known, public information about famous people such as actors, musicians, and politicians.
Although the rewritten Wikipedia biographies from Textwash may not appear verbatim in the GPT-3 training data, the model was definitely trained on the original Wikipedia entries~\cite[Section 2.2]{NEURIPS2020_1457c0d6}, as well as several other information sources about the ``anonymized'' celebrities (Table \ref{tab:emma}).
Since a vast amount of PII of public figures is already publicly available information, it is possible to re-identify famous people using the semantic context of documents that is left after PII removal \cite{lohr2024grassco}.

\begin{figure}
\centering
\includegraphics[width=0.6\linewidth]{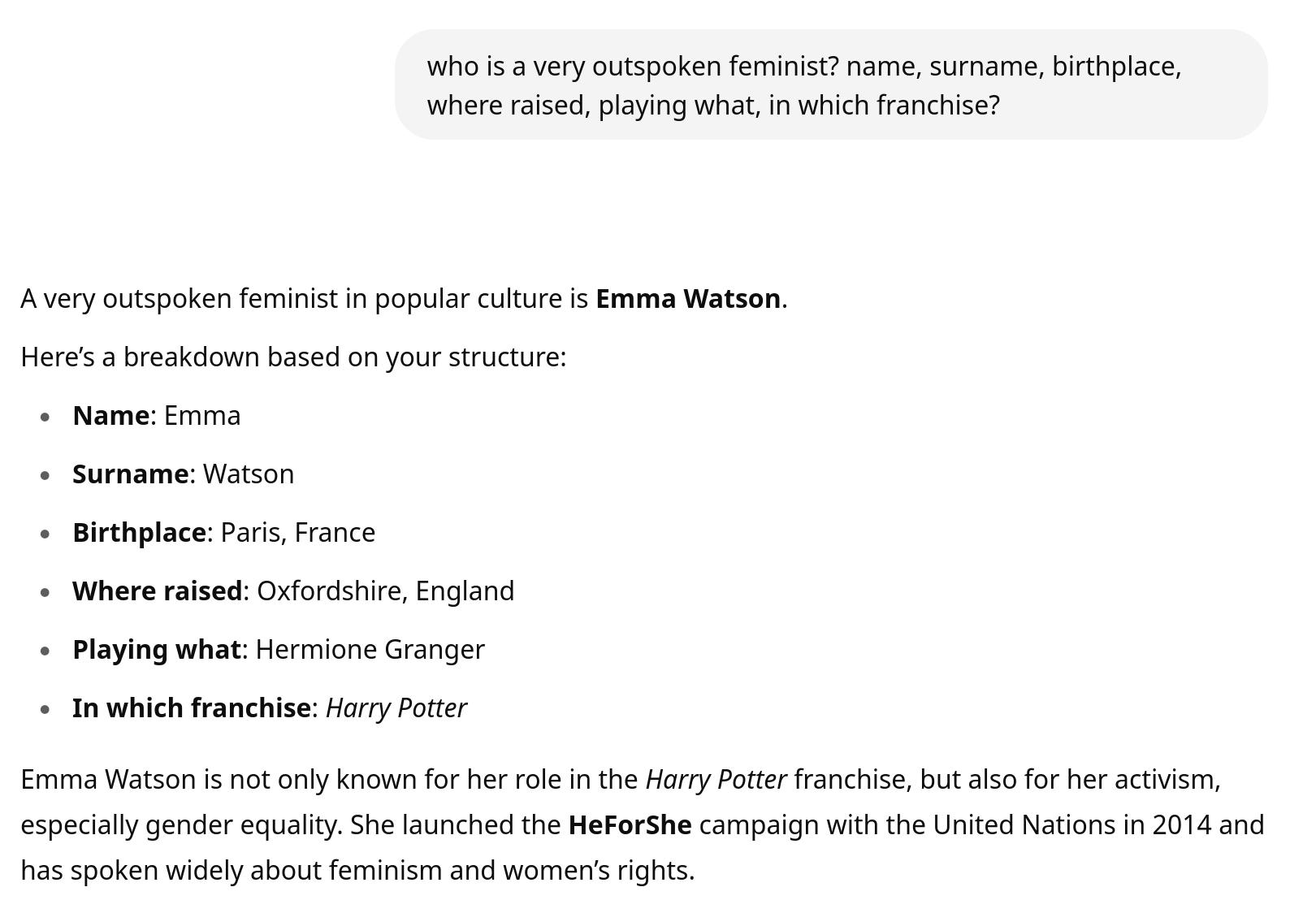}
\caption{\label{fig:emma.watson} Solving the task by \citet{Patsakis2023-de} by querying ChatGPT-4o with a few keywords only (cf.~Table \ref{tab:emma}).}
\end{figure}

We demonstrate the flaws of this benchmark in Figure~\ref{fig:emma.watson} which shows that only a few trigger words are enough to retrieve all PIIs from a language model. Moreover, Table~\ref{tab:angelina} shows another problem in the evaluation. Both solutions of the PII reconstruction are possible, as both Emma Watson and Angelina Jolie would perfectly fill the gaps. Yet the benchmark seems to be skewed towards more famous persons who might be also over-represented in the Web data fed into LLM pre-training, leading to artifacts in the benchmark and the attack evaluation. In other words, if Angelina Jolie were in the gold standard data, the PII attack filling-in Emma Watson would fail, thus lowering the attack score from which we might conclude that the PII removal technique is safer. But both for Angelina Jolie and Emma Watson, the PII removal tool would produce the \emph{very same output}.

We therefore argue that the re-identification success of GPT-3 is not the fault of the de-identification method per se, but rather a consequence of the public nature of the utilized data.
\textbf{When the original and anonymized documents share semantically similar or even identical content, recognizing the corresponding famous person becomes a pattern-matching task instead of a privacy breach.}

\begin{table*}[t]
\centering
\small
\begin{tabularx}{\textwidth}{@{}lX@{}}
\toprule
\textbf{Gold standard} & \mybox{Emma} \mybox{Watson} is a \mybox{31} year old actress, born in \mybox{Paris} but raised in the \mybox{UK}. She is most famously known for playing \mybox{Hermione} \mybox{Granger} in the \mybox{Harry Potter} franchise. [...] \mybox{she} is very involved in activism and is a very outspoken feminist.\\
\midrule
\textbf{Alternative solution} & \mybox{Angelina} \mybox{Jolie} is a \mybox{49} year old actress, born in \mybox{Los Angeles} but raised in the \mybox{Palisades, New York}. She is most famously known for playing \mybox{Lara} \mybox{Croft} in the \mybox{Lara Croft} franchise. [...] \mybox{she} is very involved in activism and is a very outspoken feminist.\\
\bottomrule
\end{tabularx}
\caption{\label{tab:angelina} Both Emma Watson and Angelina Jolie would be a valid solution of the PII reconstruction attack by \citet{Patsakis2023-de}.}
\end{table*}

\subsection{Leakage through memorization}
\label{leak-mem}
\citet{10179300} propose a PII reconstruction attack on sanitized documents.
Here, the attacker has access to an LLM fine-tuned on the original documents.
The researchers fine-tune GPT-2 models \cite{radford2019language} with and without differentially private stochastic gradient descent (DP-SGD, a training method that injects noise into gradients and therefore limits the influence of any single data sample on the model) on three datasets; ENRON \cite{10.1007/978-3-540-30115-8_22}, Yelp-HEALTH\footnote{\url{https://business.yelp.com/data/resources/open-dataset/}} and cases of the European Court of Human Rights (ECHR, \citet{chalkidis-etal-2019-neural}).
After fine-tuning, the PII in each text of the above-mentioned datasets is masked with Presidio.
The authors then attack the sanitized documents with the fine-tuned language models by filling in the masked information.
As expected, the GPT-2 models trained with privacy guarantees using DP-SGD perform worse at reconstructing the masked PII, achieving under 1\% accuracy on the ``person'' entity type on each dataset.  
In contrast, their non-private counterparts are able to guess the correct ``person'' with significantly greater accuracy, up to 15\% accuracy on ENRON and up to 18\% on the ECHR dataset, posing a greater risk to privacy.
The authors also observe that larger model versions of GPT-2 generally leak more PII when attacked with the same method.

Although \citet{10179300} claim that PII removal techniques are insufficient to prevent PII leakage, their attack does not prove that PII leak from sanitized documents because of the weak protection offered by PII removal methods.
As the attacking GPT-2 models were fine-tuned on the original data \textit{before} PII removal, the sensitive information in the data is already memorized by the model at the start of the attack and it can use the sanitized texts as context to accurately recall PII based on its training objective of next token prediction.
Consequently, the ability of the attack to reconstruct PII from sanitized texts should be attributed to \textbf{memory leakage instead of limitations in the PII removal method itself.} 
This is further proven by the authors themselves, as GPT-2 models fine-tuned on the PII removed instead of original documents achieve 0\% reconstruction accuracy.

\citet{charpentier-lison-2025-identification} face a similar issue in their approach.
The authors attack sanitized Wikipedia biographies, ECHR cases, and synthetic clinical notes by combining a retrieval model with an LLM-based infilling method to find relevant passages in background documents and infer the original masked text spans.
However, the unsanitized versions of the sanitized Wikipedia and ECHR texts are most likely in the pretraining data of their utilized infilling models, as \citet{charpentier-lison-2025-identification} acknowledge in their limitations.
The authors also vary the access of background knowledge the attacker has in their setup. 
They show in their appendix section E that an attacker without access to the original ECHR case linked to a sanitized document, but with access to all other original cases, is able to reconstruct the name of Polish government agent J. Wo{\l}\k{a}siewicz.
However, J. Wo{\l}\k{a}siewicz also appeared in other ECHR cases as Polish public official.
This does not show that the attack was successful in this case, but rather that (1) the same PII in ECHR court cases can be found in multiple documents and (2) public figures were incorrectly labeled as private entities.
Again, we argue that the supposedly successful reconstruction attack does not demonstrate that PII removal techniques are a weak privacy protection method, but that attack performances are overestimated through memorization and other factors, such as mislabeling. 

Even though \citet{morris-etal-2022-unsupervised} do not focus on PII reconstruction attacks, they encounter an interesting phenomenon in their adversarial re-identification metric. 
The authors propose an unsupervised de-identification method based on k-anonymity, where documents are continuously redacted until the probability of re-identification estimated by an adversary falls below a threshold, depending on how many people share similar attributes. 
The main de-identification target is a small subset of Wikipedia biographies.
However, the re-identification models of \citet{morris-etal-2022-unsupervised}'s method are based on RoBERTa \cite{roberta}, which was pretrained on large corpora that include Wikipedia.
The authors observe that in some cases the adversarial model is able to identify people from biographies where 95\% of the words are masked \cite[Table 5]{morris-etal-2022-unsupervised}.
While the authors state in their limitations that memorization ``is unlikely to be happening", we are not convinced by this claim, especially since the re-identification models are also fine-tuned on the biographies. 
It is highly probable that the adversarial model memorized the exact sentence structure of these individuals, which allows re-identification even when most of the text is replaced with masked tokens.
In these cases, the de-identification method has no chance to protect the privacy of the re-identified persons, as the attacker has already memorized the original biographies and their distinctive phrasing.
This is the same kind of memory leakage we also observed in \citet{10179300}'s work.

\subsection{Adversarial attacks on PII removed texts are flawed}

Overall, the estimated privacy risks posed by contemporary attacks on PII removed texts are overestimated, as either the attack models had already memorized the supposedly ``private'' data during training \cite{Patsakis2023-de, 10179300} or the ``private'' information was already publicly available in news articles \cite{nyffenegger-etal-2024-anonymity}.
While we believe that the existing attacks are valuable, we argue that data leakage is the most likely reason for the success of these adversarial attacks against sanitized documents, giving the attacker an unrealistic advantage against PII removal techniques.
Therefore, the answer to our first research question is yes: there are inherent flaws in the evaluation of current PII reconstruction attacks.

\section{The setup of adversarial attacks on PII removed texts}
\label{sec:setup}
We now address our second research question: can the flaws we identified in the previous section be resolved without access to real, sensitive data?
Given the uncertainty whether previous attacks are successful because of weaknesses in PII removal or through unintentional data leakage, we first need to outline the conditions a valid attack setup must satisfy to avoid exaggerating the attack results. 
We cover both the defense method and threat model, before turning to the data issue in Section \ref{sec:data-discussion}.

\subsection{Defense methods}
When attacking PII removal methods to discover potential weaknesses, the defensive mechanism itself has to reflect real-world deployments.
It must be highly accurate at detecting PII in texts to demonstrate its privacy protection abilities, while maintaining the text utility for downstream tasks to a reasonable degree.
Current state-of-the-art methods combine pattern matching with regular expressions and NER, though NER has significantly improved in recent years due to the integration of context-sensitive embeddings \cite{JEHANGIR2023100017}.
Therefore, the defense mechanism should at least use such a hybrid, state-of-the-art method that ideally aligns with legal frameworks, supporting GDPR or HIPAA compliance.
Furthermore, the text de-identification technique should be open-source, ensuring that its PII detection and replacement strategies are transparent and comprehensible.   
This also allows researchers to identify and address potential weaknesses in the defense mechanism.
In contrast, using closed-source, cloud-based services similar to Amazon Comprehend\footnote{\url{https://docs.aws.amazon.com/comprehend/latest/dg/pii.html}} or Microsoft Azure\footnote{\url{https://learn.microsoft.com/en-us/azure/ai-services/language-service/personally-identifiable-information/language-support?tabs=text}} to remove PII from documents would make such an error analysis more difficult.
Furthermore, if the documents contain sensitive private data, it may not be advisable to share that data with proprietary services for PII removal. 

\subsection{Threat model}
Previous attacks do not demonstrate that PII removal is ineffective for privacy protection, but rather show that PII removal cannot prevent leakage when the model already memorizes the underlying data beforehand.
However, we \emph{do} believe that PII removal methods may be vulnerable to attacks where failures can be attributed entirely to PII removal.

In recent years, the ability of LLMs to infer information from context has improved significantly, partially due to better reasoning capabilities and larger model sizes \cite{wei2022emergent}.
Furthermore, stylometric features, such as writing style, syntactic patterns and unique phrasings, can uniquely identify individuals, as famously demonstrated by the FBI in the capture of the Unabomber \cite[Section II]{leonard2017forensic}. 
When provided with free-form text written by humans, LLMs are also capable of accurately attributing authors to the corresponding documents \cite{ai-etal-2022-whodunit, huang-etal-2024-large}, which could be leveraged to infer sensitive user information.
Additionally, LLMs can be exploited to extract private information from text data that is often not explicitly stated, such as age, gender and occupation \cite{staab2024beyond}.
LLMs can also infer locations from human-written texts based on implicit geographic features, namely land marks, dialects or local cuisine \cite{zhang2025social}.  
Even the Big Five personality traits \cite{Costa2008-oz} of users can be predicted from social media posts with reasonable reliability, as demonstrated by \citet{Peters2024-fi}.
 
PII removal methods are well-suited for modifying or removing explicitly stated private information in texts. 
However, it remains uncertain how much of the data may be recovered from the implicit information left behind.
Given the capabilities of LLMs to infer various sensitive information from textual data, even when such attributes are not directly mentioned, inference attacks on PII-removed documents seem plausible with state-of-the-art reasoning models.
\textbf{However, to the best of our knowledge, inferring PII with LLMs has yet to be explored for PII removed documents while ensuring zero data leakage.}

\section{The unsolvable problem of transparent research and private data}
\label{sec:data-discussion}

While the conditions outlined in the previous section can be satisfied by researchers, the requirements on the data needed to demonstrate flaws in PII removal techniques are a different matter. 
We now examine which data sources are viable to complete our answer to the second research question.

\subsection{Public data}
There are numerous datasets publicly available when researching PII removal techniques.
In the medical domain, the MIMIC-III (Medical Information Mart for Intensive Care, \cite{mimic3}) and MIMIC-IV \cite{mimic4} datasets contain millions of free-form clinical texts, including radiology reports, patient progress notes or discharge summaries. 
Similarly, the i2b2 (Informatics for Integrating Biology and the Bedside, \cite{i2b2}) de-identification dataset also provides clinical texts, but annotated specifically for training PII removal tools.  
Several datasets from other domains offer additional resources for text sanitization.
For example, the MAPA (Multilingual Anonymisation for Public Administrations, \cite{arranz-etal-2022-mapa}) project contains both legal and medical documents in the 24 official languages of the EU.  
Furthermore, the TAB (Text Anonymization Benchmark, \cite{pilan-etal-2022-text}) corpus provides 1K annotated court cases of the ECHR and is specifically designed to evaluate text anonymization methods.    
The Big Code PII dataset \cite{bigcode} consists of 12K code samples from The Stack \cite{thestack}, covering 31 different programming languages and featuring rare PII such as IP addresses or cryptographic keys.   
Another dataset was released by \citet{Holmes2023-nz}, which comprises 22K student-written essays and includes annotations for PII to aid the development of de-identification systems. 

\paragraph{Public data are unusable for attacking PII removal}
\textbf{Although public datasets are well-suited to train PII detection models, they are inherently unsuitable to demonstrate attacks on PII removal techniques.}
Publicly available PII datasets are already de-identified by masking sensitive information or replacing it with synthetically generated data.
While this is necessary to protect the privacy of data contributors, it also removes the ground truth needed to evaluate potential attacks.
Furthermore, to prevent data leakage, one must either limit the selection of threat models to open-source LLMs trained exclusively on public data, or abandon using public sources altogether.
Any overlap between training and PII dataset will falsify the evaluation of an attack on PII removal, but their disjunction can be verified when the training data of the attack model is publicly available.
However, ``open-data'' models fall significantly behind proprietary LLMs in inference and reasoning capabilities, as demonstrated by their results on popular benchmarks \cite{hendrycks2021measuring, wang2024mmlupro, chatbotarena}.
As for closed-source LLMs, the disjunction between their training data and public PII datasets is not verifiable, potentially leading to training/test data contamination \cite{balloccu-etal-2024-leak}, and therefore they should not be combined to attack PII removal techniques.

\subsection{Why should we not use synthetic data instead?}
\label{sec:synth}
Alongside collections of human-written texts, researchers have also developed synthetically generated datasets with artificial PII to improve PII removal systems.     
The Open PII Masking 500k dataset \footnote{\url{https://huggingface.co/datasets/ai4privacy/open-pii-masking-500k-ai4privacy}}, for example, contains half a million synthetic samples from multiple languages and domains to substitute private data, generated by Llama 3.1 and 3.3 \cite{llama3}. 
SynthPAI \cite{yukhymenko2024a} is a dataset explicitly designed for personal attribute inference, providing 8K synthetic texts generated by GPT-4 to mimic the Reddit dataset published by \citet{staab2024beyond}.

In comparison to public datasets, synthetic PII datasets seem more promising for evaluating attacks.
They offer an alternative to real-world private data that does not compromise the privacy of real persons.
When created from scratch, synthetic data may enable closed-source LLMs as attack models, since the data was not part of their pretraining data.
Synthetic text generation methods also provide researchers with controls about the text domain and which PII should be included in the data, while also offering ground truth labels for evaluating PII removal attacks.
However, using synthetically generated texts to attack PII removal involves significant risks.

\paragraph{LLMs reproduce original training data}
LLMs are trained on vast amounts of text data scraped from the web, which also contains personal information \cite{kim2023propile, subramani-etal-2023-detecting}.
However, LLMs are capable of regurgitating their training data verbatim \cite{carlini2023quantifying, ippolito-etal-2023-preventing}.
When these LLMs are then asked to generate synthetic texts that resemble private data, the risk of replicating real, sensitive information from their pretraining data inherently increases.
Although guard rails reduce the risk of leaking training data, there are no formal guarantees that they hold, as it is possible to extract training data even from aligned models \cite{nasr2025scalable}.
If the synthetic texts then contain realistic PII, a successful attack against such samples can again be attributed to data leakage, due to the likely prior knowledge of the attacking LLM.
Essentially, \textbf{synthetic data generated by LLMs trained on public data scraped from the internet, is again public data and inherits the very same issues in regards to data leakage.} 

\paragraph{Differentially private synthetic texts are not a viable alternative}
To prevent private data from appearing in the synthetic texts, one can leverage differentially private methods during text generation \cite{utpala-etal-2023-locally, xie2024dp}.
While the resulting synthetic texts may suffice for downstream tasks such as text classification, their utility for mimicking private data is limited, as the text quality generally degrades, losing semantic and syntactic coherence \cite{yue-etal-2023-synthetic, meisenbacher-matthes-2024-thinking, ochs-habernal-2025-private}.
Especially in expert domains such as healthcare, text coherence is important, for example to maintain consistency in patient records.
However, recent research demonstrates that differentially private synthetic texts currently do not achieve the required quality for practical applications \cite{ramesh-etal-2024-evaluating}.
Therefore, synthetic texts generated by differentially private methods are not a suitable replacement for real private data when evaluating attacks against PII removal techniques. 

\paragraph{Biases in synthetic data}
Texts generated by LLMs can be biased towards certain features, such as gender, stereotypes \cite{kotek2023gender} or nationalities \cite{narayanan-venkit-etal-2023-nationality}, and may be limited in their expressiveness based on prompt templates and strategies \cite{chen-etal-2024-unveiling-flaws}.
Although biases per se do not disqualify the utility of synthetic data, there exists a measurable distribution shift between synthetically generated and human-written texts \cite{pilluta2021mauve, seegmiller-preum-2023-statistical}.
Synthetic data therefore reflect the complexity and diversity of human-written texts less accurately than truly sensitive datasets \cite{guo-etal-2024-curious}.
Consequently, evaluating  attacks against PII removal attacks with synthetic data becomes tricky.
On one hand, the attack model may rely on correlations between sensitive information and their surrounding text for PII recovery. 
In contrast to human-written texts, these correlations may not exist in synthetic texts, which would result in a reduced attack success rate.
Thus the PII removal technique would appear safer than it would be when applied to private documents.
On the other hand, the synthetic text generation model and the attack model may share some common biases, especially if their pretraining data overlap. 
When the generation model is asked to produce artificial names, it may produce names that appear most frequently in a given language, e.g. ``James Smith'' for U.S. English \footnote{\url{https://www.statista.com/statistics/279713/frequent-combinations-of-first-and-last-name-in-the-us/}}. 
If the attack model has the same language bias, it may also propose ``James Smith'' more often whenever it estimates that a removed span should contain a person's name, which would increase the attack success rate, even though the reconstruction may stem from the shared bias.
Successful attacks against synthetic texts after PII removal may therefore not be transferable to actual private data.

\subsection{Conundrum: Real attacks are impossible without real private data}
Adversarial attacks on PII removal techniques using synthetic or public data are inherently flawed, as they can not mitigate the risk of data leakage.
As researchers trying to disprove the effectiveness of PII removal techniques, this leaves us only with one option: evaluating attacks on private data (that was not part of the pretraining of any LLM).
Therefore, the answer to our second research question is no, as without private data, researchers cannot avoid the same problems we identified in previous attacks.

\paragraph{Access to private data clashes with requirements of transparent research}
Although private data exists, attempting to use it introduces a new set of challenges.
In 2006, \citet[Sec.~4.2]{medlock-2006-introduction}  already recognized that releasing realistic texts to benchmark PII removal techniques is impossible without prior anonymization, which, in turn, undermines the trustworthiness of the evaluation. 
However, to refute PII removal we need to identify possible ways of acquiring private data.
Therefore, we categorize three sources of real private text data: 
(1) data held by private companies, for example emails \cite{chen2019gmail};
(2) data held by public institutions, such as patient records \cite{sutton-etal-2025-named}; 
and (3) data obtained illegally through breaches or leaks, such as the Clinton email dataset.\footnote{\url{https://wikileaks.org/clinton-emails/}}

The access to the first two sources is understandably restricted for privacy protection and heavily regulated by corporate policies or national privacy acts.
Without cooperation, public researchers can not acquire private data from private or public institutions for their research.
Even when access to private data is granted, transparent research is further impeded by reasonable constraints.
For example, institutions may demand that their collected private data can not leave their premises, such as medical data stored in hospital data centers \cite{Rieke2020, lohr2024grassco} or police body camera footage \cite{voigt17police}, requiring researchers to conduct their experiments on site. 
Furthermore, there might be restrictions on how the experimental results conducted on the private data are shared with the public research community.

Instead of gaining access to private data from companies or public institutions, researchers can also use data breaches or leaks to acquire private data. 
However, research on data obtained through illicit manners raises serious ethical concerns \cite{Ienca2021}, and necessitates the approval of a research ethics board (REB).

\paragraph{Case study: ethical limitations on privacy research}
Without access to legitimate sources of private data as discussed above, we requested an approval from the research ethics board (REB) of our institution to work with leaked data containing real-world PII.
In accordance with previous research \cite{pilan-etal-2022-text, 10179300, staab2024beyond}, we argue that the common practice of removing PII from documents may be insufficient to protect the privacy of individuals. 
Our goal is therefore to disprove the effectiveness of PII removal techniques by sanitizing the leaked data with publicly available text sanitization software and develop attacks to undo the sanitization process.
We proposed to only publish the attack methodology for transparent research purposes, while not disclosing any information about the underlying text data beyond aggregate statistics such as the distribution of PII categories, and to handle the leaked text documents in a data-secure environment.

However, the REB of our institution rejected our proposal due to three major concerns.
First, conducting research on leaked personal data, whose use lacks informed consent from affected individuals, is deemed illegal for public institutions.
Second, the results of such research could lead to the creation of a harmful attack that may reconstruct PII from already published sanitized text documents, compromising the privacy of individuals who were assumed to be protected. 
Without clearly defined mitigation strategies, publishing such an attack---even without sharing the underlying data---is considered unethical.
Last and least, the REB argued that synthetic data could be a suitable alternative to leaked data, therefore avoiding the legal and ethical implications created by working with real-world PII.

Indeed, legal issues arising from using leaked data present a significant hurdle for publicly funded research, placing such data beyond our reach as public researchers.
While we argue that synthetic data is insufficient to disprove PII removal techniques (Section~\ref{sec:synth}), the second concern by the REB raises an interesting question.
Is it not our duty as privacy researchers to transparently publish the results of such an attack to demonstrate why it succeeds and therefore enabling others to develop more effective mitigation strategies?
This is common practice in the cybersecurity community, where attacks against encryption algorithms \cite{Biham1993, kocher1996, bleichenbacher1998chosen, stevens2017collision}  and communication protocols \cite{moller2014poodle, adrian2015imperfect, aviram2016drown} are published to expose vulnerabilities in existing defense mechanisms and ultimately improve the security systems we rely on. 
It is true that already published, sanitized texts may be compromised when PII reconstruction attacks are released, since they are already publicly available and cannot be retroactively protected.
However, publishing successful attacks against PII removal techniques is important to prevent the future release of even more text data relying on potentially insufficient privacy protection mechanisms.

\section{Two small-scale example studies on legal documents and videos}

As public researchers, we are unable to fulfill the strict requirements we determined that must be met to disprove PII removal as privacy protection.
Nonetheless, we run a small PII reconstruction attack under relaxed conditions.
While we can not refute PII removal this way, the study aims to contribute to the scenarios in which PII removal techniques offer insufficient privacy protection.

\subsection{Setup}
\begin{table*}[t!]
\centering
\small
\setlength{\tabcolsep}{6pt}
\begin{tabularx}{\linewidth}{@{}l *{10}{>{\raggedleft\arraybackslash}X}@{}}
\toprule
 & \textbf{Loc.} & \textbf{Org.} & \textbf{Person} & \textbf{NRP} & \textbf{Time} & \textbf{URL} & \textbf{Tel.} & \textbf{Email} & \textbf{Misc.} & \textbf{Total} \\
\midrule
\textbf{YT} & 747 & 151 & 110 & 102 & 4  & 4  & 0  & 1 & 1  & 1120 \\
\textbf{CZ} & 294 & 190 & 243 & 2   & 23 & 13 & 16 & 3 & 21 & 805 \\
\bottomrule
\end{tabularx}
\caption{\label{tab:diag} \# of PII spans by entity type in our two datasets (YT = YouTube transcripts; CZ = Czech court announcements). Abbreviations: Loc. = location, Org. = organization, NRP = nationality/religious/political group, Time = date/time, URL = web address, Tel. = telephone number, Misc. = other numerical PII, such as bank account numbers.}
\label{pii-entities}
\end{table*}

\paragraph{Data and threat model}
Instead of private data, for the first part of the study we consider public data that is unlikely to appear in web-scale crawls and consequently in pretraining corpora. 
We decide to include Czech court announcements that were published online as PDF files, announced in October 2018.
These documents are posted at the courts' web portals and the URLs to the PDFs are deleted after maximum of 30 days; the URLs are never made public again.
The data collection pipeline for a LLM data training would therefore need to discover the URLs, download the PDF files and extract their texts within a short time frame.
Though not impossible, we deem this improbable and continue under the assumption that these documents are not part of LLM training data.
Yet the PDF documents themselves had not been deleted, which might have happened by a human mistake in the court administration. We discovered the URLs of these old published PDFs by pure luck in another research project.
Our small collection consists of 28 Czech documents, which often display the personal names, birth dates and home addresses as PII of the individuals addressed by the court in the announcements.
For our experiments, we extract the text from each PDF file and translate it to English sentence-by-sentence with the \texttt{NLLB-200-3b} model \cite{nllb}. 

Our second data source might be known to an LLM model depending on the choice of the particular model.
In contrast to proprietary LLMs, open-weight LLMs have a clear training cut-off date.
Data created after the release (or the last update) of the model therefore cannot be part of the training data.
As LLM for our PII reconstruction attack, we therefore select \texttt{gpt-oss-120b} \cite{openai2025gptoss120bgptoss20bmodel}, a strong open-weight reasoning model, whose weights were published on August 5, 2025.
This allows us to create a second dataset for our experiment from videos uploaded to YouTube, utilizing their video transcripts as text data.
Using a mixture of manual selection and the platform's API, we compile a dataset of 41 transcripts that meet the following criteria: 
(1) the videos are travel video blogs (vlogs), as creators may mention their name, locations they are visiting or other personal details that qualify as PII; 
(2) they were uploaded between August 6th and October 5th 2025 (our collection date) to avoid training data overlap with \texttt{gpt-oss-120b}; 
(3) the videos are longer than three minutes, to exclude short-form content and obtain longer transcripts;
(4) the view count of each video is under 100 (at the time of collection), which may reduce the chance that transcripts from prior uploads on the same channel are part of \texttt{gpt-oss-120b}'s pretraining data; 
and (5) finally, the video transcript is available in English, either manually created or automatically generated.

\paragraph{Defense and attack methodology}
\begin{table*}[t]
\centering
\small
\begin{tabularx}{\textwidth}{@{}X@{}}
\toprule
\textbf{Model prompt}\\
\midrule
You are a privacy risk evaluator.

\textbf{Input:} \\
A text containing masked spans like \texttt{[MASK\_0]}, \texttt{[MASK\_1]}, \dots{} (any number; indices may be non-contiguous).

\textbf{Task A --- Hint audit (leakage check):} \\
Scan the text for unmasked information that could help infer the masked spans.
Hints include: unmasked proper nouns, partial names, unique locations/landmarks, organization names, routes, flight/train identifiers, exact dates, distinctive numbers, usernames, emails, phone numbers, or any unusually specific phrase.
Quote each hint exactly as it appears in the text (verbatim substring).
If there are no clear leaks, still list weaker contextual hints, but label them as \texttt{"contextual"} rather than \texttt{"direct leak"}.

\textbf{Task B --- Reconstruction:} \\
Identify every unique mask token in the text.
For each mask, silently derive constraints from local context + the hints you listed.
Produce EXACTLY 3 candidate strings per mask (ranked: most likely first).

\textbf{Rules:} \\
Do NOT output your reasoning steps.
Do NOT add any keys beyond the required JSON schema.
Guesses must be only the missing span text (no extra words).
Match capitalization/punctuation implied by context.
Use outside knowledge only if strongly implied by hints in the text.

\textbf{Output JSON only, with this exact schema:} \\
\texttt{\{}\\
\texttt{\hspace*{1em}"hints": [}\\
\texttt{\hspace*{2em}\{}\\
\texttt{\hspace*{3em}"text": "<verbatim hint substring>"}\\
\texttt{\hspace*{2em}\}}\\
\texttt{\hspace*{1em}],}\\
\texttt{\hspace*{1em}"guesses": \{}\\
\texttt{\hspace*{2em}"MASK\_0": ["...", "...", "..."],}\\
\texttt{\hspace*{2em}"MASK\_1": ["...", "...", "..."]}\\
\texttt{\hspace*{1em}\}}\\
\texttt{\}}\\
\textbf{Text:} \\
\bottomrule
\end{tabularx}
\caption{\label{tab:modelprompt} Prompt used for the PII reconstruction attack.}
\end{table*}

We detect and mask PII from both datasets, using the pattern-based recognizers from Presidio for structured PII and spaCy's transformer-based English model \footnote{\url{https://spacy.io/models/en\#en\_core_web\_trf}} for NER.
We additionally decide to mask organizations, as they often contain location names (e.g. District Court Prague). 
We provide an overview of the number of detected PII entities in both datasets in Table~\ref{pii-entities}.
Overall, the most frequently detected types of PII are locations, organizations, personal names, and NRP entities (nationalities, religious or political group).

After masking the PII in both datasets, we attempt to reconstruct the removed PII with \texttt{gpt-oss-120b} in the following manner:
The attack model is given each masked document individually and is prompted to act as a privacy-risk evaluator performing two tasks.
First, extract contextual clues from the text that may help to infer the masked spans.
Second, for each mask token in the document, propose three reconstruction candidates based on the collected clues.
We present our model prompt verbatim in Table~\ref{tab:modelprompt}.
For each mask token, we consider a reconstruction as successful only if any of its respective proposed candidates is an exact string match (EM) to the original text of the removed PII span.

\subsection{Results}
\begin{table*}[t!]
\centering
\setlength{\tabcolsep}{6pt}
\begin{tabularx}{\textwidth}{@{}l >{\raggedleft\arraybackslash}X | >{\raggedleft\arraybackslash}X *{5}{>{\raggedleft\arraybackslash}X}@{}}
\toprule
 & \textbf{Total} & \textbf{MaxDoc} & \textbf{Loc.} & \textbf{Org.} & \textbf{Person} & \textbf{NRP} & \textbf{Time} \\
\midrule
\textbf{YT} & 0.191 & 0.515 & 0.194 & 0.139 & 0.155 & 0.294 & 0.250 \\
\textbf{CZ} & 0.055 & 0.364 & 0.119 & 0.037 & 0.008 & 0.000      & 0.000    \\
\bottomrule
\end{tabularx}
\caption{\label{tab:attack-em3}Top-3 exact match rate (EM@3) by PII category and dataset. \textbf{Total} displays EM@3 over all masked PII within a dataset. \textbf{MaxDoc} denotes the maximum document-level EM@3 observed in each dataset. We omit four categories (URL, Tel., Email, Misc.), as they have EM@3$=0$ in both datasets.}
\label{attack-results}
\end{table*}

We present the results of our attack in Table~\ref{attack-results}.
The attack model successfully reconstructs the exact PII span within its three guesses (EM@3) in ca. 19\% of cases for the YouTube transcripts (YT) and 5.5\% of cases for the translated Czech court announcements (CZ). 
The reconstruction rate varies greatly between documents and PII categories.
While 11 of the 28 CZ documents and 10 of the 41 YT transcripts show zero exact matches, the maximum document-level EM@3 reaches ca. 36\% for CZ and ca. 51.5\% for YT, showing that some documents are more vulnerable to the attack than others.
In the YT dataset, NRP entities have the highest EM@3 rate with ca. 29\%, although personal names, locations and organizations are also repeatedly reconstructed. 
In comparison, the CZ dataset shows overall lower EM@3 scores, with considerable reconstruction rates for locations and organizations.

\subsection{Analysis}

\begin{table*}[t]
\centering
\small
\setlength{\tabcolsep}{6pt}

\begin{tabularx}{\textwidth}{@{}X@{}}
\toprule
\textbf{Reconstruction example from YT } \\
\midrule
\footnotesize
Hi everyone! \texttt{[MASK\_0]} here Today, we’re heading to \texttt{[MASK\_1]} for a 3-day trip with our friend visiting from \texttt{[MASK\_2]} The
hotel is spacious with a kitchen, and the kids are so excited At \texttt{[MASK\_3]}, we even spotted an ad for the Demon Slayer
movie - so exciting! We’ll also visit some iconic \texttt{[MASK\_4]} spots - \texttt{[MASK\_5]}, the I Love NY gift shop, and more! And now,
our Day 1 adventure begins! I’m three years old Wow, beautiful! We are here, mama! We are in \texttt{[MASK\_6]}! We are not in
\texttt{[MASK\_7]} That’s a huge bed! No more monkeys jumping on the bed Don’t do that. \texttt{[MASK\_8]} is going to be sad This is a
great location near \texttt{[MASK\_9]} Mama, there is a phone! \texttt{[MASK\_10]} likes that Yes, it’s wonderful Be gentle please Good
night Good morning! After a short rest, we walked to \texttt{[MASK\_11]} - about 20 minutes from our hotel Mail man! Look! (The
bird) had a bath I’m a birdie What’s that? \texttt{[MASK\_12]}’s getting a piggyback ride Be careful, pigeon It’s a place that
truly represents \texttt{[MASK\_13]} - filled with giant billboards and crowds of people everywhere Look! That’s mama! Is that
me?! There are amazing people They look cool, huh She said hi to me That’s wonderful Next stop - the \texttt{[MASK\_14]} store! It
smells sweet here like chocolate One two, one two Do you want to do this? Which one do you want? Do you like this one?
Not too much, OK? Don’t eat it yet It’s a little bit rainy! That’s right Then we stopped by the I Love NY gift shop Are
they friends? Yeah! There there We stopped by \texttt{[MASK\_15]} to grab dinner You are tired, aren’t you? No, I’m not sleepy
Really? Beautiful! Thank you for watching until the end! If you enjoyed this video, don’t forget to like and subscribe!
And stay tuned for Day 2 of our \texttt{[MASK\_16]} adventure Vlog! Thank you for watching! \\
\medskip
\textbf{Hints:} I Love NY gift shop; Demon Slayer movie; giant billboards and crowds of people everywhere; sweet here like chocolate \\
\bottomrule
\end{tabularx}

\vspace{1ex}

\begin{tabularx}{\textwidth}{@{}l l l X@{}}
\toprule
\textbf{Mask} & \textbf{Type} & \textbf{Gold} & \textbf{Top-3 guesses} \\
\midrule
\texttt{[MASK\_0]}  & PERSON       & \textit{person 1}           & Mom; Emily; Sarah \\
\texttt{[MASK\_1]}  & LOCATION     & New York               & New York City; NYC; Manhattan \\
\texttt{[MASK\_2]}  & LOCATION     & Japan                  & Japan; South Korea; Australia \\
\texttt{[MASK\_3]}  & LOCATION     & Times Square           & noon; afternoon; morning \\
\texttt{[MASK\_4]}  & LOCATION     & New York               & NYC; New York; Manhattan \\
\texttt{[MASK\_5]}  & ORGANIZATION & M\&M’s World            & Times Square; Central Park; Rockefeller Center \\
\texttt{[MASK\_6]}  & LOCATION     & Japan                  & Manhattan; Queens; Brooklyn \\
\texttt{[MASK\_7]}  & LOCATION     & Japan                  & Queens; Brooklyn; Bronx \\
\texttt{[MASK\_8]}  & PERSON       & \textit{person 2}                  & Mom; Dad; Teacher \\
\texttt{[MASK\_9]}  & LOCATION     & Grand Central Station  & Central Park; Times Square; Brooklyn Bridge \\
\texttt{[MASK\_10]} & PERSON       & \textit{person 2}                & Baby; Kid; Child \\
\texttt{[MASK\_11]} & LOCATION     & Times Square           & Central Park; Times Square; Brooklyn Bridge \\
\texttt{[MASK\_12]} & PERSON       & \textit{person 3}               & Kid; Baby; Girl \\
\texttt{[MASK\_13]} & LOCATION     & New York               & New York; NYC; Manhattan \\
\texttt{[MASK\_14]} & ORGANIZATION & M\&M’s                 & M\&M's World; Godiva; Lindt \\
\texttt{[MASK\_15]} & ORGANIZATION & Whole Foods            & Katz's Delicatessen; Joe's Pizza; Chinese restaurant \\
\texttt{[MASK\_16]} & LOCATION     & New York               & NYC; New York; Manhattan \\
\bottomrule
\end{tabularx}

\caption{Example masked document and extracted hints (top) and gold spans with the attack model's top-3 reconstruction candidates (bottom). Person names were manually replaced with italic placeholders.}
\label{tab:attack-example-yt}
\end{table*}

\begin{table*}[t]
\centering
\small
\setlength{\tabcolsep}{6pt}

\begin{tabularx}{\textwidth}{@{}X@{}}
\toprule
\textbf{Reconstruction example from CZ} \\
\midrule
\footnotesize
Posted on: October 30, 2018 Deadline for removal: November 30, 2018 Removed on: \textit{file number 1} Notice for
posting on the court's official notice board pursuant to Section 49(4) of the Code of Civil Procedure. Addressee:
\texttt{[MASK\_0]}, born \texttt{[MASK\_1]} Address to which the document is to be delivered: \texttt{[MASK\_2]} \textit{number}, \textit{number}
\texttt{[MASK\_3]} \textit{number} - \textit{location} Court that submitted the document for delivery: \texttt{[MASK\_4]} for \texttt{[MASK\_5]} 3
Document to be delivered: \textit{file number} - decision no. \textit{number}, lh. \texttt{[MASK\_6]} As the addressee was not present at the time
of delivery of the court document and as it was not possible to deliver the court document to a person authorized to accept it,
the court document was deposited. As the addressee did not collect the document within 10 days, the tenth day of this period is
considered the date of delivery according to the law. Since after the expiry of this period it was not possible to place the
document in the addressee's mailbox or other mailbox used by the addressee, it was returned to the sending court. You can collect
the stored shipment at the information center of \texttt{[MASK\_7]}, \texttt{[MASK\_8]}, \texttt{[MASK\_9]}, during the following hours:
Mon from 8:00 a.m. to 11:00 a.m. and from 12:00 p.m. to 4:30 p.m. Tue from 8:00 a.m. to 11:00 a.m. and from 12:00 p.m. to 4:00 p.m.
Wed from 8:00 a.m. to 11:00 a.m. and from 12:00 p.m. to 5:00 p.m. Thu from 8:00 a.m. to 11:00 a.m. and from 12:00 p.m. to 4:00 p.m.
Friday from 8:00 a.m. to 11:00 a.m. and from 12:00 p.m. to 2:30 p.m. Please bring your ID card or other proof of identity with you.
Name and surname of the court delivery person: \texttt{[MASK\_10]}, \texttt{[MASK\_11]} Date: October 30, 2018 Signature of court delivery agent \\
\medskip
\textbf{Hints:} Posted on: October 30, 2018; Deadline for removal: November 30, 2018; File number;  \textit{location};
information center of; Mon from 8:00 a.m. \\
\bottomrule
\end{tabularx}

\vspace{1ex}

\begin{tabularx}{\textwidth}{@{}l l l X@{}}
\toprule
\textbf{Mask} & \textbf{Type} & \textbf{Gold} & \textbf{Top-3 guesses} \\
\midrule
\texttt{[MASK\_0]}  & PERSON       & \textit{person 1}                 & Jan Novák; Petra Svobodová; Martin Dvořák \\
\texttt{[MASK\_1]}  & DATE\_TIME    & \textit{birth date}             & 12.05.1975; 03.07.1980; 25.12.1965 \\
\texttt{[MASK\_2]}  & LOCATION     & \textit{location}                      & Plynárenská; Kollárova; Jiráskovo \\
\texttt{[MASK\_3]}  & LOCATION     & Prague                          & Praha; Prague; Praha \\
\texttt{[MASK\_4]}  & ORGANIZATION & District Court                  & District Court; Municipal Court; Regional Court \\
\texttt{[MASK\_5]}  & LOCATION     & Prague                          & civil; criminal; administrative \\
\texttt{[MASK\_6]}  & DATE\_TIME    & 10/11                           & 12; 13; 14 \\
\texttt{[MASK\_7]}  & ORGANIZATION & the District Court for Prague 3 & Prague; Praha; Czech Republic \\
\texttt{[MASK\_8]}  & LOCATION     & Jagellonská 5                   & Municipal; District; Regional \\
\texttt{[MASK\_9]}  & LOCATION     & Prague 3                        & Court; Court of Justice; Court \\
\texttt{[MASK\_10]} & PERSON       & \textit{person 2}                 & Pavel; Jiří; Martin \\
\texttt{[MASK\_11]} & PERSON       & \textit{person 3}                     & Kučera; Novák; Horák \\
\bottomrule
\end{tabularx}

\caption{Example masked court announcement (top), extracted hints, and gold spans with the attack model's top-3 reconstruction candidates (bottom). Italic text was manually removed to protect the privacy of individuals before publication.}
\label{tab:attack-example-cz}
\end{table*}

Despite applying PII masking to the YT and CZ documents, our attack resulted in an alarming EM@3 rate. 
How can this happen?
We identify a key contributor through the generated \textit{hints} produced by our attack: incomplete PII detection. 
For example, the successful reconstruction of personal names in the CZ dataset was a result of repeated mentions within a single document. 
The sentence ``\textit{Person name} confirms that this is a true copy of the original.'' appeared three times, but Presidio failed to flag the name in one instance, which then enabled the attack model to infer the other masked instances.
We observe similar patterns when analyzing the hints generated for the YouTube transcripts. 
As shown in Table~\ref{tab:attack-example-yt}, leaving ``I Love NY gift shop'' and ``giant billboards and crowds of people everywhere'' unmasked is sufficient for the attack model to guess that the vlog was likely filmed in New York and its famous Times Square.
This leads to a domino effect, as once the PII detection system misses a single entity, the attack model may correctly recognize the unmasked text as relevant and produce guesses based on it for other masks.
This not only leads to an increased reconstruction rate for locations, but other entity types, too, as the nationality of persons in a text can be estimated based on the country, leading to the high EM@3 rate for NRP entities.
Unfortunately, automatic PII removal systems are not 100\% reliable, as they currently do not achieve a 100\% recall rate for PII in texts. 
Therefore, before releasing documents processed with automatic PII removal techniques, a human review is necessary to identify and remove missed PII manually.

Additonally, the high reconstruction rate for personal names in the YT transcripts is mostly based on guessing the names of public or religious figures, such as ``Monty Python'', ```Murasaki Shikibu'', ``Buddha'' or``god'', but the attack fails to recover the names of the vlog creators when they introduce themselves.
When detecting PII entities in texts, Presidio does not differentiate between private and public information, and consequently removes both. 
This can inflate the attack's success rate, as many reconstructions are based on publicly known entities rather than lesser-known or private ones.

We also observe that, when the surrounding context provides little to no information, the attack model tends to make generic guesses.
Specifically in the CZ dataset, when a person's name is masked, the model defaults to generic suggestions for the most common Czech names, such as ``Jan Novák; Petra Svobodová; Martin Dvořák'', as shown for \texttt{[MASK\_0]} in Table~\ref{tab:attack-example-cz}.
While we could not find any of those names in the CZ dataset, an attack model may successfully reconstruct PII based on generic guesses, even if the context offers no information for reliable inference. If you were ``Jan Novák'', under the current PII attack formalization your privacy risk would be higher, which is paradoxical given that you have a very generic name.

\subsection{Limitations}
In the YouTube transcripts, creators often describe what they see.
As our dataset consists of travel vlog transcripts, these descriptions often include popular travel destinations.
However, similar descriptions of these distinctive places are widely available online, for example through blog entries, travel guides or even Wikipedia articles. 
Therefore, the pretraining data of LLMs already include information about these places, which enables LLMs to infer the location from context even though the names of the locations are removed from documents.
This highlights that even though the attack model has not been trained on the YT transcripts, we could not avoid leakage, as the content of the YT dataset overlaps with texts and world knowledge the attack model has already acquired during pretraining.   
As discussed in Section~\ref{leak-mem}, PII removal techniques can not prevent such inferences, as the removed information is already public information.

The reconstruction rates on the CZ dataset may be influenced by the sentence-level translation to English, which can alter the context.
Additionally, some entities, such as personal names and parts of addresses, remained in Czech after translation, which potentially reduced the PII detection accuracy of the English Presidio pipeline. 
Furthermore, even though we consider it unlikely, our attack model may have already seen the Czech court announcements during pretraining.

Another limitation of our attack setup is the reliance on a single prompt. 
While the attack is simple and reproducible, differently formulated, more sophisticated prompts may achieve higher EM@3 rates as our approach.

\section{Conclusion}
PII removal techniques are commonly used to remove private information from texts to comply with data privacy regulations and protect the privacy of the individual.
Fundamentally, these methods mimic what people did in the 1950s to remove classified information from documents by manually blacking out sensitive text spans.
While this approach may suffice to hide information from other humans without contextual knowledge of a sanitized document, PII removal techniques cannot offer any formal privacy guarantees. 

Therefore, it is easy to assume that attacks against PII removal are successful given its inherent flaws.
However, analyzing existing attacks leads us to question this assumption:
Is the evaluation of existing methods for PII reconstruction attacks fundamentally flawed?
Our answer is yes: 
The success rate of previous attacks is overestimated, as data leakage is not excluded as an impact factor from the experimental setups.

But is it possible for public researchers to address this issue without access to real, sensitive data?
Despite our analysis of previous attacks, there is substantial reason to believe that adversarial attacks against PII removal techniques may successfully infer PII when solely providing the PII removed documents, without relying on data leakage as we demonstrated in the analysis of our small-scale experiments.
However, determining whether an adversarial attack truly compromises PII removed texts requires its evaluation on sensitive, private data, because public and synthetic data do not sufficiently prevent data leakage. 
Unfortunately access to private data is severely limited for public researchers due to strict privacy regulations, institutional policies, and ethical considerations.
Therefore, as members of the public research community, \emph{we currently cannot attack PII-removed documents in a transparent, reproducible, and trustworthy manner.}

\paragraph{The way forward}
Despite the seemingly unsolvable problem of empirical research on attacks on PII removal methods, we believe there is an alternative route. We observe that in existing attacks against PII removal techniques, it can not be excluded that the attacker already knew the private information before the attack, or the supposedly private information to protect was already publicly available. 
Future research should therefore strive to design properly formalized threat models, inspired by research in cybersecurity or differential privacy, and to mathematically define the entire life cycle including defense mechanisms, data access assumptions, capabilities of the attacker, and alike. We believe that having such a formal framework, or even multiple formal frameworks with different sets of assumptions and axioms, would allow us to exactly reason about the individual components, to properly evaluate them, and to properly communicate the privacy protection and privacy risks. To the best of our knowledge, none of the existing attack or privacy formalisms currently fits the big picture of text anonymization and LLM PII attacks on datasets of unknown ownership status.

Unfortunately a new axiomatic model would inevitable lead to a new theory of privacy. As such, it is not a bad goal to pursue, as we are not aware of any formal model that can faithfully capture the world we live in---the textual datasets we share privately or publicly, the LLMs having access to things we do not know what they have access to such as proprietary models like Gemini possibly trained on the entire Google ecosystem, the publicly unknown data available to secret services or OSINT as open-source intelligence, and alike---none of that fits any current framework well. Research on formal privacy guarantees seems to predominantly adopt differential privacy, but DP makes very specific probabilistic claims whose semantic is easy to be misunderstood \citep{Kifer.et.al.2022.arXiv} and also does not really model the process of creating the datasets in the first place \citep{Kifer.Machanavajjhala.2011.PODS}. But in the scenarios we explore in this paper, the problem is exactly data exchange and flow, and who got some information about whom (recall the Swiss court case in Section~\ref{sec:leakage.through.media.reporting}), for which DP is not the right framework. The same problem pertains to the alternatives of DP, including $k$-anonymity or all other dozens of methods from Chapter~2 of \citet{Fung.et.al.2011.book}. Recently there has been some renaissance of the Contextual Integrity (CI) model, where information flow is modeled explicitly between agents and their roles in contexts \citep{liPrivacyChecklistPrivacy2025}, but the formal adoption of CI is not principled, it is rather ad-hoc through prompts and in combination with legal frameworks such as the GDPR. True formalization of CI was attempted already by \citet{Barth.et.al.2006.SP} by using temporal logic, but the gap between this formalization and anything usable as an underlying framework for PII an anonymization is too wide. To summarize, we do see the need for a principled theoretical framework for PII removal and privacy protection in the era of LLMs, which would help us prevent hidden experimental flaws we report in this paper, but proposing one is beyond the scope of this current paper and is left for future work.

\section*{Acknowledgements}

This work has been partly supported by the Research Center Trustworthy Data Science and Security (\url{https://rc-trust.ai}), one of the Research Alliance centers within the \url{https://uaruhr.de} and by the German Federal Ministry of Education and Research and the Hessian Ministry of Higher Education, Research, Science and the Arts within their joint support of the National Research Center for Applied Cybersecurity ATHENE.

\bibliographystyle{compling}
\bibliography{anthology, custom}

\end{document}